%% file: emnlp2021.tex
\documentclass[11pt]{article}

\usepackage[]{emnlp2021}

\usepackage{times}
\usepackage{latexsym}
\usepackage{amsthm,amsmath,amssymb}
\usepackage{mathrsfs}
\usepackage{algorithm}  
\usepackage{algorithmicx}  
\usepackage{algpseudocode}

\usepackage[T1]{fontenc}

\usepackage[utf8]{inputenc}

\usepackage{microtype}
\usepackage{graphicx, subfigure}

\title{Evaluating the Capability of Large-scale Language Models \\ on Chinese Grammatical Error Correction Task}

\author{Fanyi Qu \quad Chenming Tang \quad Yunfang Wu\thanks{ \; Corresponding author} \\
  National Key Laboratory for Multimedia Information Processing, Peking University    \\
  MOE Key Laboratory of Computational Linguistics, Peking University  \\
  School of Computer Science, Peking University \\
  \texttt{\{fanyiqu@, tangchenming@stu, wuyf@\}pku.edu.cn} \\}

\begin{document}
\maketitle
\begin{abstract}
Large-scale language models (LLMs) have shown remarkable capability in various of Natural Language Processing (NLP) tasks and attracted lots of attention recently. However, some studies indicated that large language models fail to achieve promising result beyond the state-of-the-art models in English grammatical error correction (GEC) tasks. In this report, we aim to explore how LLMs perform on Chinese GEC (CGEC) tasks and provide guidance for future work. We conduct experiments with 12 LLMs of different model scales on 4 Chinese GEC datasets. Our experimental results indicate that the performances of LLMs on automatic evaluation metrics (e.g. $F_{\text{0.5}}$ score) falls short of the previous state-of-the-art (SOTA) models because of the problem of over-correction. Furthermore, we also discover notable variations in the performance of LLMs when evaluated on different data distributions and the priority of general-purpose models over their reasoning counterparts. Our findings demonstrates that further investigation is required for the application of LLMs on CGEC task.
\end{abstract}

\section{Introduction}

Building on InstructGPT~\cite{DBLP:conf/nips/Ouyang0JAWMZASR22}, ChatGPT has demonstrated its powerful ability to understand complex instructions and generate reasonable responses on various of NLP tasks. Following the technical trajectory of ChatGPT, a significant number of high-quality LLMs have emerged in recent times in both academia and industry, such as LLaMA~\cite{DBLP:journals/corr/abs-2302-13971}, ChatGLM~\cite{du2022glm} and PaLM~\cite{DBLP:journals/corr/abs-2305-10403}. Previous studies found that these LLMs have achieved great performance on a wide range of NLP tasks, including machine translation~\cite{DBLP:journals/corr/abs-2301-08745}, named entity recognition~\cite{DBLP:journals/corr/abs-2305-05862}, text summarization~\cite{DBLP:journals/corr/abs-2302-08081}, etc.

Certain studies have taken comprehensive investigations into the performance of LLMs in the domain of English GEC, yielding some interesting findings~\cite{DBLP:journals/corr/abs-2304-01746,DBLP:journals/corr/abs-2303-13648}. LLMs are not able to outperform SOTA models in terms of automatic evaluation metrics. This is primarily because LLMs tend to make unnecessary modifications to make the input sentences more fluent, which may result in over-correction problem, and in some cases, even alter the original semantics of the input sentences. 

In this report, we aim to explore the performance of LLMs in Chinese GEC (CGEC) task. We conducted experiments on various LLMs to investigate the influence of model size on the GEC results. Additionally, we attempted different test dataset from various data sources to explore the impact of data distribution on the outcomes. Experimental results indicate that up-to-date LLMs still lag behind SOTA CGEC models. We also find that model performance is subject to data distribution and general-purpose language models outperform powerful reasoning models even at lower cost.

\section{Experimental Setup}

\begin{table*}[]
    \centering
    \begin{tabular}{c|cccccc}
        \hline
        Dataset & Data Source & \#Sents & \#Err. Sent. & Avg. Length & Avg. Edits & Avg. Refs  \\
        \hline
        NLPCC & Chinese learner & 2000 & 1983 & 29.7 & 2.1 & 1.1 \\
        MuCGEC & Chinese learner & 1136 & 1136 & 44.0 & 4.0 & 2.2 \\
        \hline
        FCGEC & Native speaker & 2000 & 1101 & 55.8 & 1.0 & 1.3 \\
        NaCGEC & Native speaker & 500 & 482 & 56.2 & 1.2 & 1.2 \\
    \hline        
    \end{tabular}
    \caption{Statistics of datasets.}
    \label{tab:dataset}
\end{table*}

\begin{figure*}[htbp]
\centering
\includegraphics[width=2\columnwidth]{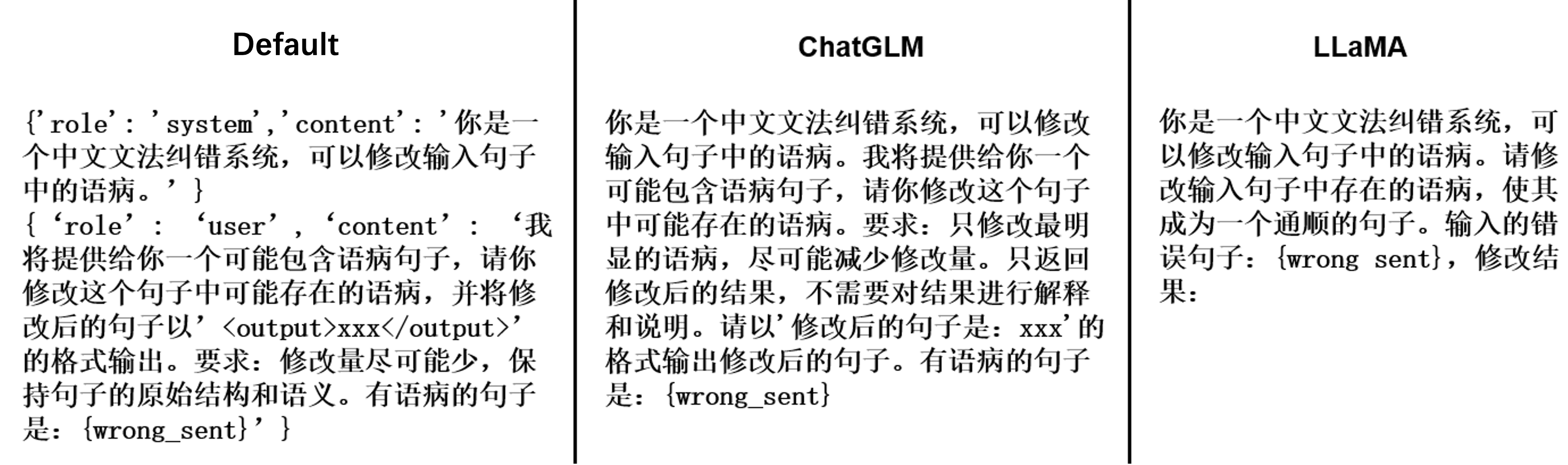}
\caption{The prompt for LLMs. ChatGLM-6B and LLaMA-7B are prompted with specially designed prompts. Other models use the default one.}
\label{fig:prompt}
\end{figure*}

\subsection{Dataset}
We conduct experiments on four CGEC datasets to provide a comprehension demonstration of LLMs' capability. The detailed statistics of these datasets are shown in Table~\ref{tab:dataset}.

\subsubsection{GEC data from Chinese learners}
We apply the test set of NLPCC-2018~\cite{DBLP:conf/nlpcc/ZhaoJS018} and the validation set of MuCGEC~\cite{DBLP:conf/naacl/0004LBLZLHZ22} for evaluation. These two datasets collect the grammatical errors made by foreigners during their process of learning Chinese.

\subsubsection{GEC data from Chinese native speaker examination}
We apply the validation set of FCGEC~\cite{DBLP:conf/emnlp/XuWPFC22} and the validation set of NaCGEC~\cite{ma2022linguistic} for evaluation. These two datasets are collected from Chinese native speakers' language examinations.

\subsection{Model}
We conduct experiments on 12 LLMs with different model scales:
\begin{itemize}
    \item OpenAI's ChatGPT Herd of Models~\footnote{https://platform.openai.com/docs/api-reference}: we evaluate the performance of \texttt{gpt-3.5-turbo}, \texttt{gpt-4o-mini} and \texttt{o3-mini} with OpenAI's API. The first two are general-purpose models while the third is a powerful reasoning model.
    \item ByteDance's Doubao~\footnote{https://www.doubao.com}: we evaluate \texttt{doubao-1.5-pro} with Volcengine's API~\footnote{https://www.volcengine.com/docs/82379}.
    \item DeepSeek Herd of Models~\cite{DeepSeekV3, DeepSeekR1}: we evaluate \texttt{deepseek-v3} and \texttt{deepseek-r1} with Volcengine's API. The former is a general-purpose model while the latter is a powerful reasoning model.
    \item ChatGLM Herd of Models~\cite{du2022glm, ChatGLM}: we evaluate \texttt{chatglm-6b} on 4 NVIDIA 3080Ti GPUs and \texttt{glm3-130b} with Volcengine's API.
    \item LLaMA Herd of Models~\cite{DBLP:journals/corr/abs-2302-13971}: we evaluate \texttt{llama-7B} on 4 NVIDIA 3080Ti GPUs and \texttt{llama-3.1-8b-instruct} on one NVIDIA A40 GPU.
    \item Qwen Herd of Models~\cite{Qwen25}: we evaluate \texttt{qwen-2.5-7b-instruct} on one NVIDIA A40 GPU and \texttt{qwq-32b} on one NVIDIA A100 GPU. The former is a general-purpose language model and the latter is a reasoning model.
\end{itemize}

\subsection{Evaluation Metric}
We evaluate models' performance with Precision (P), Recall (R) and $F_{\text{0.5}}$ from word level and char level respectively.

We adopt the official implementation of MaxMatch ($M^2$)~\cite{DBLP:conf/naacl/DahlmeierN12} scorer to calculate word-level $F_\text{0.5}$ score and choose PKUNLP as our word segment tool. We apply ChERRANT~\footnote{https://github.com/HillZhang1999/MuCGEC/tree/main/

scorers/ChERRANT} for char-level metric calculation.

\input{charlevel}

\subsection{Prompt}
Considering the differences in performance of large language models, we designed different prompts for them. These prompts are roughly the same in semantics, but there are some differences in details. The prompts are shown in Figure~\ref{fig:prompt}.


\input{wordlevel}

\section{Experimental results}

Char-level and word-level experimental results are shown in Table~\ref{tab:char-level} and \ref{tab:word-level} respectively.

First, different data sources result in distinct evaluation results. Most LLMs exhibit significantly superior performance on Chinese learner data (NLPCC and MuCGEC), as opposed to Chinese native speaker examination data (FCGEC and NaCGEC). According to our observations, the grammatical errors made by Chinese learners primarily involve the misuse of similar words or phrases, rather than incorrect sentence structures. In contrast, GEC data from Chinese native speaker examination maintains a higher level of regularity and is consisted of more complex structural errors. 
It is noteworthy that there exist gaps between GEC data from Chinese examination and Chinese native speakers' daily spoken habits. Specially, Doubao, QwQ and the DeepSeek models demonstrate competitive performance on Chinese native speaker examination data. This might be due to data leakage or specialized GEC content in training data.

Second, newer models perform better than prior ones and larger models are better than smaller ones. For example, the newer LLaMA-3.1 significantly outperforms the older LLaMA on all datasets, while larger Doubao and DeepSeek outperforms all other relatively smaller models.

Third, general-purpose models are better than reasoning models at CGEC. The up-to-date reasoning models \texttt{o3-mini} and \texttt{deepseek-r1}, while costing more money and time, fails to outperform their relatively older counterparts \texttt{gpt-4o-mini} and \texttt{deepseek-v3}. The newer and larger \texttt{qwq-32b}, albeit performing better than \texttt{qwen-2.5-7b-instruct}, is at the cost of 4 times larger model size and much longer inference time (64 hours on a single NVIDIA A100 GPU while \texttt{qwen-2.5-7b-instruct} only takes 1 hour on a single NVIDIA A40). This indicate that the long deep thinking process of reasoning models cannot contribute to their CGEC performance. In practical applications, it is recommended to apply general-purpose language models for CGEC task.

Fourth, there still exist great gaps between SOTA models and LLMs on automated evaluation metrics. Previous work~\cite{DBLP:journals/corr/abs-2301-08745} has found the problem of over-correction for LLMs, which has also been noticed in our experiment. 

What's more, it is hard to explain why the char-level evaluation metrics is significantly lower than word-level evaluation metrics, which is not noticed in previous work. 

\section{Conclusion}
In this report, we explore the performance of various LLMs on CGEC task. Experimental results indicate that there still remains gap between LLMs' performance and current SOTA models. Furthermore, the performance of different LLMs' performance is subject to the distribution of test data and general-purpose models are better at CGEC than reasoning models at lower cost. Future work can focus on addressing the over-correction problem of LLMs and explore the untapped potential of LLMs in the field of GEC tasks.

\bibliography{anthology,custom}
\bibliographystyle{acl_natbib}

\end{document}

%% file: charlevel.tex
\begin{table*}[htbp]
    \centering
    \small
    \tabcolsep=1.3mm
    \begin{tabular}{c|ccc|ccc|ccc|ccc}
        \hline
        \hline
         & \multicolumn{3}{|c|}{\textbf{NLPCC}} & \multicolumn{3}{|c|}{\textbf{MuCGEC}}  & \multicolumn{3}{|c|}{\textbf{FCGEC}} & \multicolumn{3}{|c}{\textbf{NaCGEC}} \\
          & P & R & $F_\text{0.5}$ & P & R & $F_\text{0.5}$ & P & R & $F_\text{0.5}$ & P & R & $F_\text{0.5}$ \\
        \hline
        GPT-3.5-turbo & 18.07 & 26.80 & 19.33 & 22.79 & 30.03 & 23.94 & 3.38 & 12.88 & 3.96 & 5.38 & 12.30 & 6.06 \\
        GPT-4o-mini & 22.55 & 30.93 & 23.85 & 21.78 & 23.93 & 22.18 & 9.50 & 16.11 & 10.35 & 9.51 & 16.97 & 10.43 \\
        o3-mini & 21.29 & 31.93 & 22.81 & 20.56 & 25.19 & 21.35 & 8.24 & 19.44 & 9.31 & 8.59 & 21.66 & 9.77 \\
        Doubao-1.5-pro & 30.19 & 39.96 & \underline{31.74} & 27.46 & 29.58 & \textbf{27.86} & 32.18 & 45.08 & \textbf{34.14} & 35.93 & 51.53 & \textbf{38.25} \\
        DeepSeek-V3 & 31.05 & 37.12 & \textbf{32.10} & 25.66 & 25.62 & \underline{25.65} & 22.63 & 29.80 & \underline{23.78} & 23.77 & 32.31 & \underline{25.10} \\
        DeepSeek-R1 & 20.27 & 35.39 & 22.16 & 15.88 & 25.23 & 17.15 & 18.78 & 36.06 & 20.77 & 19.31 & 37.30 & 21.37 \\
        ChatGLM-6B & 16.76 & 7.54 & 13.47 & 22.93 & 7.45 & 16.20 & 4.36 & 4.96 & 4.47 & 8.41 & 5.07 & 7.43 \\
        GLM3-130B & 16.56 & 31.27 & 18.28 & 16.16 & 23.63 & 17.25 & 11.59 & 26.84 & 13.07 & 12.07 & 28.42 & 13.64 \\
        LLaMA-7B & 8.90 & 5.43 & 7.89 & 13.37 & 7.04 & 11.34 & 1.43 & 1.78 & 1.49 & 2.83 & 2.17 & 2.67 \\
        LLaMA-3.1-8B-Instruct & 16.30 & 20.06 & 16.94 & 19.58 & 16.54 & 18.89 & 6.35 & 9.25 & 6.78 & 5.37 & 8.86 & 5.83 \\
        Qwen-2.5-7B-Instruct & 19.51 & 27.43 & 20.70 & 20.84 & 21.53 & 20.98 & 11.98 & 19.56 & 12.98 & 11.75 & 19.86 & 12.80 \\
        QwQ-32B & 20.82 & 31.58 & 22.34 & 18.08 & 21.67 & 18.70 & 17.33 & 32.15 & 19.09 & 17.55 & 33.21 & 19.38 \\
        \hline
        SOTA & - & - & - & 72.27 & 21.69 & 49.62 & 59.21 & 41.57 & 54.58 & 52.83 & 40.43 & 49.78 \\
        \hline
    \hline        
    \end{tabular}
    \caption{Experimental results (char-level). The first and second highest $F_\text{0.5}$ scores are in \textbf{bold} and \underline{underlined} respectively.}
    \label{tab:char-level}
\end{table*}

%% file: wordlevel.tex
\begin{table*}[htbp]
    \centering
    \small
    \tabcolsep=1.3mm
    \begin{tabular}{c|ccc|ccc|ccc|ccc}
        \hline
        \hline
         & \multicolumn{3}{|c|}{\textbf{NLPCC}} & \multicolumn{3}{|c|}{\textbf{MuCGEC}}  & \multicolumn{3}{|c|}{\textbf{FCGEC}} & \multicolumn{3}{|c}{\textbf{NaCGEC}} \\
          & P & R & $F_\text{0.5}$ & P & R & $F_\text{0.5}$ & P & R & $F_\text{0.5}$ & P & R & $F_\text{0.5}$ \\
        \hline
        GPT-3.5-turbo & 28.69 & 33.92 & 29.60 & 36.69 & 36.26 & \underline{36.61} & 5.47 & 14.94 & 6.26 & 10.25 & 17.83 & 11.20 \\
        GPT-4o-mini & 32.12 & 37.78 & 33.11 & 32.84 & 29.32 & 32.07 & 12.24 & 17.33 & 13.00 & 13.01 & 20.84 & 14.06 \\
        o3-mini & 32.15 & 40.27 & 33.50 & 33.30 & 32.19 & 33.07 & 12.78 & 23.42 & 14.05 & 13.58 & 26.97 & 15.08 \\
        Doubao-1.5-pro & 39.07 & 46.86 & \underline{40.41} & 37.42 & 34.31 & \textbf{36.75} & 35.37 & 46.73 & \textbf{37.18} & 39.44 & 56.27 & \textbf{41.95} \\
        DeepSeek-V3 & 40.35 & 43.21 & \textbf{40.89} & 37.82 & 31.20 & 36.28 & 25.82 & 30.76 & \underline{26.67} & 26.85 & 35.61 & \underline{28.24} \\
        DeepSeek-R1 & 33.49 & 45.79 & 35.38 & 32.14 & 35.13 & 32.70 & 24.36 & 38.51 & 26.29 & 24.77 & 42.11 & 26.99 \\
        ChatGLM-6B & 24.60 & 9.82 & 18.91 & 34.31 & 9.38 & 22.41 & 7.09 & 5.65 & 6.74 & 16.98 & 7.89 & 13.80 \\
        GLM3-130B & 29.44 & 41.81 & 31.29 & 30.35 & 32.63 & 30.78 & 17.13 & 30.51 & 18.78 & 18.14 & 34.50 & 20.04 \\
        LLaMA-7B & 17.63 & 7.98 & 14.20 & 24.54 & 10.02 & 19.03 & 3.42 & 3.00 & 3.33 & 8.44 & 4.75 & 7.30 \\
        LLaMA-3.1-8B-Instruct & 26.10 & 26.82 & 26.24 & 29.74 & 20.76 & 27.38 & 9.85 & 11.63 & 10.17 & 10.53 & 13.86 & 11.06 \\
        Qwen-2.5-7B-Instruct & 29.26 & 34.59 & 30.19 & 31.53 & 26.10 & 30.27 & 16.35 & 22.04 & 17.24 & 16.08 & 23.73 & 17.18 \\
        QwQ-32B & 32.39 & 40.30 & 33.71 & 32.14 & 29.94 & 31.68 & 22.66 & 34.50 & 24.33 & 23.19 & 38.11 & 25.16 \\
        \hline
        SOTA & 57.10 & 28.90 & 47.80 & - & - & - & - & - & - & - & - & - \\
        \hline
    \hline        
    \end{tabular}
    \caption{Experimental results (word-level). The first and second highest $F_\text{0.5}$ scores are in \textbf{bold} and \underline{underlined} respectively.}
    \label{tab:word-level}
\end{table*}